\documentclass{ieeeaccess}
\usepackage{cite}
\usepackage{amsmath,amssymb,amsfonts}
\usepackage{algorithmic}
\usepackage{graphicx}
\usepackage{textcomp}
\usepackage{float}
\usepackage{hyperref}

\usepackage{bm}
\makeatletter
\AtBeginDocument{\DeclareMathVersion{bold}
\SetSymbolFont{operators}{bold}{T1}{times}{b}{n}
\SetSymbolFont{NewLetters}{bold}{T1}{times}{b}{it}
\SetMathAlphabet{\mathrm}{bold}{T1}{times}{b}{n}
\SetMathAlphabet{\mathit}{bold}{T1}{times}{b}{it}
\SetMathAlphabet{\mathbf}{bold}{T1}{times}{b}{n}
\SetMathAlphabet{\mathtt}{bold}{OT1}{pcr}{b}{n}
\SetSymbolFont{symbols}{bold}{OMS}{cmsy}{b}{n}
\renewcommand\boldmath{\@nomath\boldmath\mathversion{bold}}}
\makeatother

\def\BibTeX{{\rm B\kern-.05em{\sc i\kern-.025em b}\kern-.08em
    T\kern-.1667em\lower.7ex\hbox{E}\kern-.125emX}}

\begin{document}
\history{Date of publication xxxx 00, 0000, date of current version xxxx 00, 0000.}
\doi{10.1109/ACCESS.2024.0429000}

\title{Medalyze: Lightweight Medical Report Summarization Application Using FLAN-T5-Large}
\author{\uppercase{Van-Tinh Nguyen}\authorrefmark{1}, \IEEEmembership{Member, IEEE},
\uppercase{Hoang-Duong Pham}\authorrefmark{2},
\uppercase{Thanh-Hai To}\authorrefmark{2},
\uppercase{Cong-Tuan Hung DO}\authorrefmark{2},
\uppercase{Thi-Thu Trang Dong}\authorrefmark{3}, 
\uppercase{ Vu-Trung Duong Le}\authorrefmark{4}, and \uppercase{Van-Phuc Hoang}
\authorrefmark{1},
\IEEEmembership{Member, IEEE}}

\address[1]{Electrical Engineering Department, Le Quy Don Technical University, Ha Noi, Viet Nam ( take@lqdtu.edu.vn)}
\address[2]{Department of Information Technology, University of Science and Technology of Hanoi, Ha Noi, Viet Nam}
\address[3]{Department of acute diseases and
Emergency - Institute for the treatment of senior Staff, 108 Institute of
Clinical Medical and Pharmaceutical Sciences, Ha Noi, Viet Nam }
\address[4]{Computing Architecture Laboratory, Nara Institute of Science and Technology, Nara, Japan}

\markboth
{Author \headeretal: Preparation of Papers for IEEE TRANSACTIONS and JOURNALS}
{Author \headeretal: Preparation of Papers for IEEE TRANSACTIONS and JOURNALS}

\corresp{Corresponding author: Van-Phuc Hoang (phuchv@lqdtu.edu.vn).}

\begin{abstract}

Understanding medical texts presents significant challenges for both healthcare professionals and patients due to the complexity of terminology and the context-dependent nature of medical language. To address this issue, this paper introduces Medalyze, an AI-powered application designed to enhance the comprehension of medical texts through advanced natural language processing (NLP) techniques. Medalyze leverages three specialized Flan-T5-Large models, each fine-tuned for a distinct task: (1) summarizing key information from medical reports, (2) extracting health-related issues from conversational exchanges, and (3) identifying the primary question within a given medical text.
The application offers a multi-platform solution, featuring both a web-based interface and a mobile application, ensuring accessibility for a diverse user base. Data management is powered by YugabyteDB, enabling seamless synchronization across platforms. To ensure efficient performance, Medalyze is deployed using a scalable API architecture, providing real-time text processing. A comprehensive evaluation was conducted to assess the performance of Medalyze against GPT-4, demonstrating superior performance in specialized medical text summarization tasks. User feedback indicates significant improvements in understanding medical passages, with enhanced clarity and usability.
Medalyze represents a practical and efficient tool for improving communication and decision-making in healthcare settings, bridging the gap between complex medical language and user-friendly information for both professionals and non-specialist audiences.
\end{abstract}


\titlepgskip=-21pt

\maketitle

\section{Introduction}
\label{sec:introduction}
\PARstart{M}{edical} transcriptions are essential for healthcare, providing detailed information about patient conditions, diagnostic results, and treatment plans \cite{b1}--\cite{b3}.
However, the inherent complexity of medical terminology and the extensive nature of these documents present significant challenges \cite{b4}--\cite{b6}. Medical professionals often face difficulties in efficiently identifying key insights within lengthy reports, while patients and caregivers may struggle to understand specialized language and technical details, hindering informed decision-making and proactive health management. The rapid digitization of healthcare records, coupled with the increasing adoption of AI-driven tools, underscores the need for solutions that enhance the comprehension of medical content \cite{b7}--\cite{b15}. As a response to this demand, Medalyze is introduced as an AI-powered application designed to bridge the communication gap between healthcare providers and non-specialist audiences. By leveraging cutting-edge natural language processing (NLP) techniques, the application facilitates the interpretation of medical texts without compromising accuracy, thereby enhancing accessibility for a diverse range of users. Medalyze not only aids healthcare professionals by reducing the cognitive load associated with interpreting medical transcriptions but also empowers patients by simplifying complex information. This paper provides a comprehensive analysis of the technical development, key features, and performance evaluation of Medalyze, highlighting its potential to transform the way medical content is accessed and understood. By addressing key problems in healthcare communication, Medalyze aims to improve decision-making, enhance patient outcomes, and foster a more inclusive healthcare ecosystem.\\
The main contributions of this study are given as:
\begin{itemize}
    \item Introducing Medalyze, an AI-powered application leveraging three specialized Flan-T5-Large models, each tailored to perform distinct tasks:
        \begin{enumerate}
            \item Summarizing key information in medical reports (text format);
            \item Extracting health issues from conversational exchanges; and
            \item Identifying the central question within a given passage.
        \end{enumerate}
    \item Establishing a multi-platform ecosystem for Medalyze, featuring a web-based interface (developed using HTML, CSS, and JavaScript) and a mobile application for Android (built using Java). The underlying data management is powered by YugabyteDB, enabling efficient data synchronization and sharing across platforms. APIs ensure smooth communication between system components, enhancing the usability and reliability of the application.
    \item Conducting a performance analysis of Medalyze within the healthcare ecosystem, with a comparative evaluation against GPT-4.
\end{itemize}

The remainder of this paper is organized as follows: The related works are presented in Section II. The background and research methodology used to build Medalyze are presented in Sections III and IV. Next, Sections V and VI cover the source of dataset acquisition, data processing, and pre-trained models. Section VII introduces the evaluation metrics, while Section VIII details the web and mobile application architecture. Section IX provides a comparative analysis with GPT-4. Finally, Section X concludes the paper.  

\section{Related works}

The use of AI and NLP in processing medical texts has seen significant growth in recent years, with numerous projects and research initiatives aiming to enhance healthcare communication \cite{b16}--\cite{b22}. One prominent example is the application of advanced AI models like ChatGPT, Claude, and Gemini, which have been explored for their ability to alleviate clinical documentation burdens by processing and summarizing complex clinical texts \cite{b23}--\cite{b35}. These models demonstrate the potential of AI to streamline workflows and improve the accessibility of medical content.

Another significant advancement involves the adoption of FLAN-T5 models, which have been applied to tasks such as transforming clinical text into structured Fast Healthcare Interoperability Resources (FHIR) formats and summarizing conversations to enhance information accessibility \cite{b36}--\cite{b38}. These models highlight the versatility of fine-tuned language models in addressing a wide range of medical text processing requirements.\\
Efforts by organizations like Dataloop have resulted in the development of specialized AI tools capable of summarizing complex medical documents, research papers, and clinical notes into concise, comprehensible formats. These solutions utilize pre-trained transformer models and extensive medical literature to enhance their efficacy, making them valuable assets in the medical field. For instance, a study conducted at Stanford University \cite{b38} investigated the use of large language models (LLMs) for summarizing clinical text. Notably, the use of Flan-T5-XL in this study demonstrated competitive performance despite having a relatively low parameter count of 2.7 billion, compared to models like Alpaca, Vicuna, Llama-2, and even GPT-4. This competitive performance is attributed to the model's architecture, specifically its Sequence-to-Sequence (Seq2Seq) design, which is inherently well-suited for text summarization tasks. This result was a key motivation for selecting Flan-T5-Large in Medalyze, ensuring a lightweight application with quantifiable accuracy and contextual consistency.

Public reception of these technologies has been largely positive, with studies showing that adapted LLMs can outperform humans in certain summarization tasks \cite{b39}--\cite{b50}. This demonstrates their potential to significantly enhance efficiency in clinical workflows, reduce cognitive burdens on medical professionals, and empower patients with clearer, more actionable medical information. However, challenges such as ensuring data privacy, maintaining accuracy, and addressing ethical concerns remain central to the ongoing development and adoption of these tools.

Medalyze builds upon these advancements, distinguishing itself by integrating three specialized AI models for distinct tasks: summarizing medical reports, extracting health issues from conversations, and identifying key questions within passages. This modular and multi-platform architecture ensures flexibility and accessibility, catering to a wide range of user groups, including healthcare professionals, patients, and caregivers. By prioritizing real-world usability and accuracy, Medalyze seeks to address the limitations of existing solutions, making a meaningful contribution to the evolving landscape of AI-powered healthcare tools.
\section{Backgrounds}
\begin{figure*}[h]
    \centerline{\includegraphics[scale=.55]{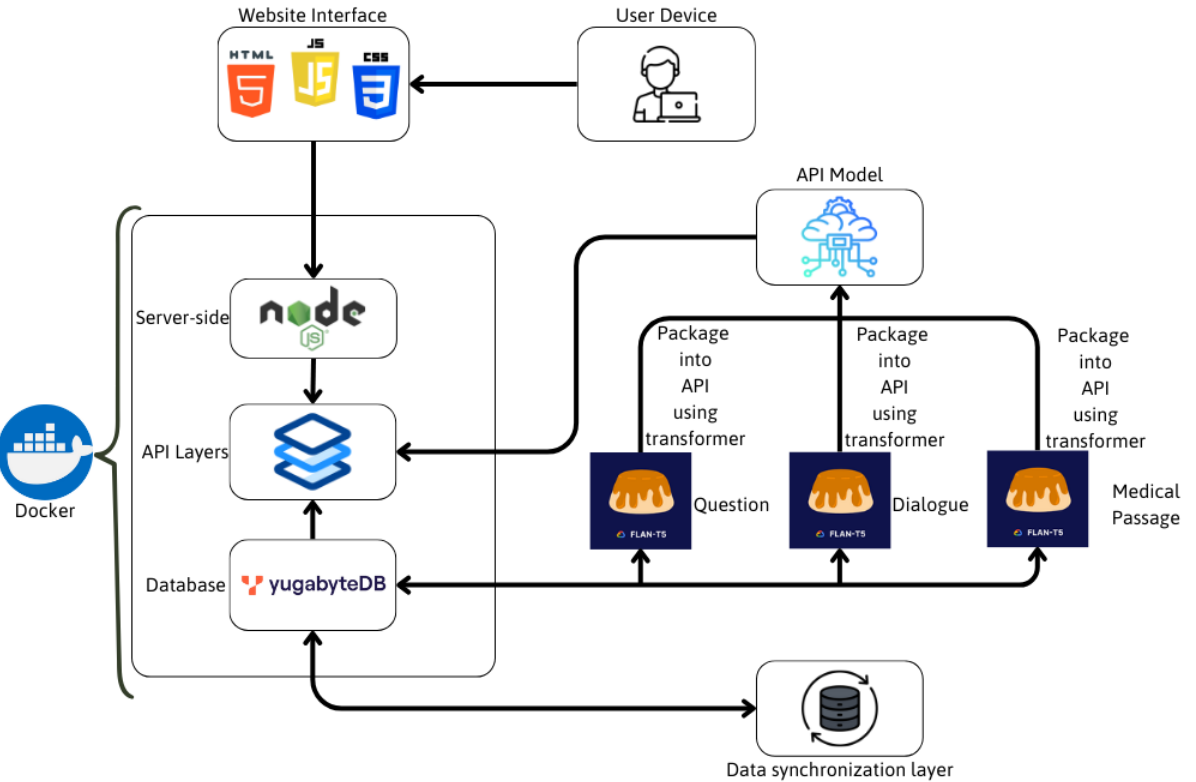}}
    \caption{System Architecture of Medalyze}
    \label{fig:System component}
\end{figure*}
The LLMs utilized in Medalyze are trained using an extensive and carefully curated dataset, ensuring they possess the depth and versatility necessary to meet diverse application needs. These models leverage state-of-the-art training techniques, including transfer learning and fine-tuning on medical-specific corpora, to enhance their ability to effectively handle complex terminology and context-dependent language.\\
To achieve seamless integration across both web and mobile platforms, the models are deployed via scalable APIs, ensuring consistent and reliable functionality across devices. This design enables real-time processing and efficient interactions between the front-end interfaces and the underlying AI models. Furthermore, the application architecture incorporates failover mechanisms and optimized response handling, ensuring high availability and strong performance.

To support this architecture, a centralized database, YugabyteDB, has been established, serving two primary functions: (1) Data Storage: Securely storing application content and user-generated summaries; (2) Data Synchronization: Facilitating smooth data exchange between the web and mobile platforms. The database is designed to efficiently handle distributed workloads, ensuring low latency and high throughput for user queries. Advanced data replication strategies are employed to maintain data integrity and consistency across platforms. Additionally, robust data encryption protocols are implemented to protect sensitive medical information during storage and transmission.

This infallible infrastructure ensures seamless data exchange and maintains coherence, providing users with a unified and efficient experience across their chosen platform. The system's modular design further allows for future scalability, enabling the integration of additional features as user needs and technical advancements evolve. To implement Medalyze, three specialized AI models were trained locally on a curated medical dataset. A rigorous filtering process was applied to ensure data quality, relevance, and compliance with project requirements. Training was conducted on a high-performance computational setup to maximize model efficiency. Key aspects of the training process include:

\begin{itemize}
    \item \textbf{Model Selection:} Flan-T5-Large was chosen for its lightweight architecture and strong performance in text summarization tasks.
    \item \textbf{Data Curation}: The dataset was carefully filtered to retain only medical texts relevant to the application's objectives.
    \item \textbf{Hyperparameter Optimization:} Key training parameters were fine-tuned to balance model performance and computation efficiency.
\end{itemize}

The system’s backend relies on YugabyteDB, deployed using Docker containers for enhanced portability and scalability. This setup supports efficient data storage, synchronization, and retrieval. Each interaction is assigned a unique identifier, enabling secure data handling. The database schema is optimized for large volumes of structured and unstructured data, supporting the system's scalability as it grows.
The front-end of Medalyze is designed to provide an intuitive user experience across both web and mobile platforms:

    \begin{itemize}
        \item \textbf{Web Application:} Developed using HTML, CSS, and JavaScript, ensuring responsive design and intuitive navigation.
        \item \textbf{Mobile Application:} Built for Android using Java, maintaining consistent functionality with the web version.
        \item \textbf{API Communication:} APIs bridge communication between the front-end interfaces and the AI models, enabling seamless data exchange.
    \end{itemize}
\section{Research Method}

Medalyze is a locally bound application designed to \textbf{summarize} medical-related text. Its development follows a structured pipeline that begins with data set acquisition, pre-processing, and model training. The trained models are integrated into a web-based interface, allowing users to input medical inquiries in text format and receive summarized responses based on the selected model.
This section outlines the technical approach used to build Medalyze, from data preparation to application deployment. Fig. \ref{fig:System component} provides an overview of the system architecture, illustrating its key components and interactions. The system consists of a web interface where users submit medical-related text queries, which are processed by fine-tuned Flan-T5-Large models packaged and integrated using Flask. The selected model generates a summarized output, which is then returned to the user. The backend manages model selection, data processing, and response generation, while a database stores relevant information for future retrieval.
\begin{figure}[tb]
    \centering
    \includegraphics[width=0.88\linewidth]{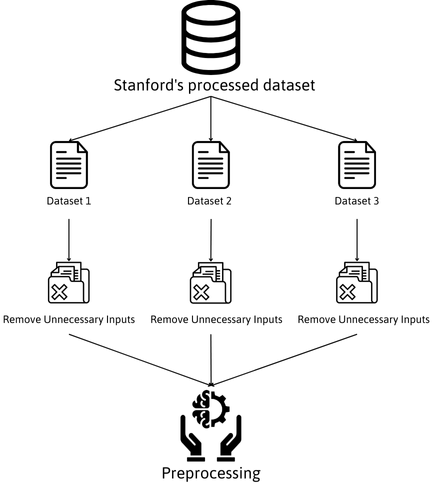}
    \caption{Diagram of Data Source Preparation for Model Training}
    \label{fig:2}
\end{figure}

\section{Dataset acquisition}
The dataset used for training the models is derived from a study conducted by a research team at \textbf{Stanford University}. Their research has been publicly shared, and the dataset is available as an open-source resource in their GitHub repository \cite{b38}. The dataset is provided in JSON, where each entry consists of ID, inputs, and target. The same format is used for the Flan-T5-Large models as well.

To acquire the necessary dataset for this paper, the first step involved extracting Stanford’s publicly available dataset for thorough analysis. Stanford’s research team trained multiple models, including but not limited to Alpaca, Med-Alpaca, and Vicuna, using a single dataset. 
A preliminary assessment was performed to categorize the dataset based on model requirements. The data was then organized into distinct subsets according to content type (e.g., conversational exchanges vs. structured medical passages) to align with the objectives of the three Flan-T5-Large models as shown in Fig. \ref{fig:2}. Filtering criteria were established to retain only relevant medical text for further preprocessing.

In order to effectively separate a dataset, the purpose of each model has to be clear, which results in the idea behind the Flan-T5-Large models. To ensure relevance, a filtering process is applied, isolating only the subset pertaining to the models. Further refinements were applied through an automated filtering script to enhance dataset quality, ensuring compatibility with model-specific requirements. 
\begin{figure}[t]
    \centering
    \includegraphics[width=0.9\linewidth]{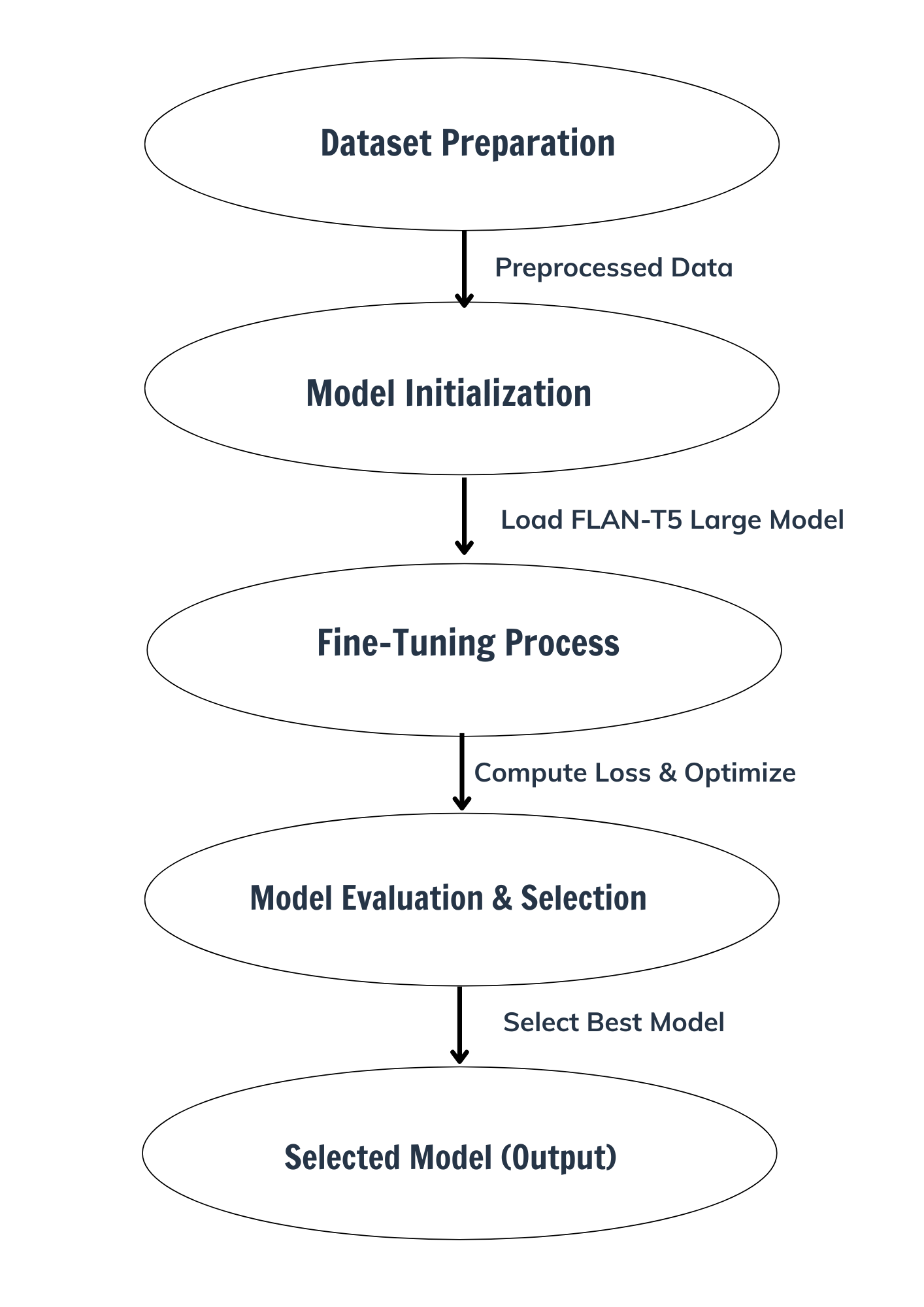}
    \caption{Workflow of Model Training and Fine-Tuning}
    \label{fig:Model workflowx}
\end{figure}
\section{Pre-train models}
Flan-T5, specifically Flan-T5-Large, is the chosen model for this paper
due to it’s lightweight as well as it following the Sequence-to-Sequence principle which is fitting for text summarizations. Before deployment, the AI models must be pre-trained on the refined dataset to optimize their summarization capabilities. The pre-training process ensures that the models effectively generate concise and relevant summaries from the given medical text.
The following features cover key aspects of model pre-training:\\
   \indent \textbf{Hardware specifications}: Determines the progress of the training process by putting the limit of memory and GPU. \\  
   \indent\textbf{Hyper-parameters adjustment}: Optimizing training parameters to balance accuracy and efficiency.\\  
   \indent \textbf{Model evaluation}: Performance evaluation using appropriate metrics to ensure reliability.\\
To effectively train the summarization models, a structured workflow is followed. Fig.\ref{fig:Model workflowx} illustrates the overall process, starting from the input of the refined data to model fine-tuning and evaluation. The workflow consists of the following key stages:\\
    \indent\textbf{Data preparation}: The dataset is prepared and ready to be fed into the training pipeline.\\    
    \indent \textbf{Model initialization}: The AI models are pre-trained with default configurations.\\  
    \indent\textbf{Fine-turning process}: The model is trained on the dataset in multiple iterations, with constant adjustment to hyper-parameters.\\
    \indent\textbf{Model evaluation and selection}: The models are evaluated to assesses their performances. \\  
    \indent\textbf{Output models}: Once optimal performance is achieved, the three models, each for their respective purposes as mentioned earlier, are saved and stored for deployment. 

\subsection{Hardware specifications}
Although hardware is not the primary factor in model training, it significantly impacts training speed. The hardware specifications used for this project are as follows:
\begin{itemize}
    \item \textbf{CPU}: Intel(R) Core i5-12400F @ 2.50GHz, 6 cores and 12 threads
    \item \textbf{GPU}: NVIDIA Quadro RTX 5000 (230W)
    \item \textbf{RAM}: 16 GB
\end{itemize}
These components are not ideal for model training, particularly due to the limited 16 GB RAM. However, effective hyper-parameter tuning allowed successful model training despite hardware limitations.
\subsection{Hyper-parameters adjustment}
This part details how the three Flan-T5-Large models were trained, with a focus on hyper-parameter configurations. While hyper-parameters were fine-tuned based on model performance, Hardware limitations also influenced hyper-parameter selection. Weaker hardware affects training speed and batch size, which will be elaborated on shortly. 
The key hyper-parameters adjusted during training include:\\
    \indent \textbf{Learning rate}: Controls how much model weights changes. The standard value is $1e^{-4}$, which was adjusted to $2e^-5$. However, after a some observation, instabilities were seen throughout the training process. Ultimately, it was settled at $1e^-5$ for a stable training as well as prevent overshooting. \\  
    \indent\textbf{Batch size}: Number of processed samples before updating weights. The Batch Size was reduced to 2, attuning to the hardware used to train the models. However, while lower value means less memory usage, it increases noise. Thus Gradient Accumulation was used.  \\
    \indent\textbf{Gradient Accumulation (steps = 2)}: Accumulate gradients from small batches before updating weights. It helps simulates larger batch size (reduce noise) keeping the memory usage low. Gradient Accumulation is often paired with low batch size to balance the noise and memory usage.  \\  
    \indent\textbf{Evaluation steps (steps = 500)}: Defines how often the model evaluates performance during the training process. The number of steps were increased to better the monitoring process of the models performance, albeit slower training. Frequent evaluations are crucial, especially in the early training stages. If there exist sign of overfitting, underfitting, or loss not improving then the training process can be stopped early to adjust hyper-parameters accordingly. \\
    \indent \textbf{Float16 (FP16)}: Uses 16-bit instead of 32-bit (or 64-bit) to save memory and speed up the training process. \\
    \indent \textbf{Early stopping}: Stops training if the performance does not improve after a set amount of steps. In addition, it prevents overfitting and save time. \\
    \indent \textbf{Weight decay}: A regularization technique is used to prevent overfitting by penalizing large weights during training. The penalty value is set to 0.01, meaning a 1\% penalty is applied to each weight update. This value is appropriate for this specific training process, as it is neither too low, which could lead to overfitting, nor too high, which could cause underfitting.
\section{Model evaluation}
Evaluating model performance is a crucial step in ensuring its effectiveness in summarizing medical text. Various evaluation metrics are employed to quantify how well the generated summaries align with human-written references. This section introduces the key evaluation metrics used in this study, including \textbf{BLEU}, \textbf{ROUGE-L}, \textbf{BERTScore}, and \textbf{SpaCy Similarity}.\\
    \indent \textbf{BLEU}: Stands for Bilingual Evaluation Understudy, which measure the similarity between the generated summary to the original text based on n-gram overlap are as follows: \\
    \begin{equation}
        \text{BLEU} = BP \times \exp\left(\sum_{n=1}^{N} w_n \log p_n \right)
        .\label{eq1}\end{equation}
        where: \\
        $p_n$: precision for n-grams \\
        $w_n$: weight for each n-gram \\
        BP: brevity penalty to penalize short translations:\\ 
        \begin{equation}
        BP =\begin{cases} 1, & \text{if } c > r \\e^{(1 - r/c)}, & \text{if } c \leq r\end{cases}
         .\label{eq2}\end{equation}
        where: \\
        $c$: length of candidate translation\\
        $r$: length of reference translation (closest match)

    \indent \textbf{ROUGE-L}:  Stands for Recall-Oriented Understudy for Gisting Evaluation. ROUGE calculates how much the of the original text is covered in the generated text. The L variant measures based on the longest common subsequence (LCS) are as follows: 
     \begin{equation}
        \text{ROUGE-L} = F_{\beta} = \frac{(1 + \beta^2) \times P_{\text{LCS}} \times R_{\text{LCS}}}{P_{\text{LCS}} + \beta^2 R_{\text{LCS}}}
         .\label{eq3}\end{equation}
        where: \\
        LCS: is the longest sequence of words appearing in both the candidate and reference.\\
         \begin{equation}
         P_{\text{LCS}} = \frac{LCS\ length}{candidate\ length}(precision)\\
        .\label{eq4}\end{equation}
          \begin{equation}
         R_{\text{LCS}} = \frac{LCS\ length}{reference\ length}(recall)\\
         .\label{eq5}\end{equation}
        $\beta$: weight balancing recall and precision

    \indent\textbf{BERTScore}: Uses embedding (BERT) to compare the generated and referenced summaries at the semantic level are as follows: 
     \begin{equation}
        \text{Precision} = \frac{1}{|X|} \sum_{x \in X} \max_{y \in Y} \cos(\text{BERT}(x), \text{BERT}(y))
     .\label{eq6}\end{equation}
         \begin{equation}
        \text{Recall} = \frac{1}{|Y|} \sum_{y \in Y} \max_{x \in X} \cos(\text{BERT}(x), \text{BERT}(y))
    .\label{eq7}\end{equation}
  
         \begin{equation}
        \text{BERTScore} = F_1 = 2 \times \frac{\text{Precision} \times \text{Recall}}{\text{Precision} + \text{Recall}}
          .\label{eq8}\end{equation}
        $X$: candidate sentence\\
        $Y$: reference sentence\\
        BERT(x): BERT embedding of token x \\
    \indent \textbf{SpaCy Similarity}:  Measures similarity between the generated and referenced summaries based on word embeddings from SpaCy are as follows: \\
        \begin{equation}
        \text{similarity}(A, B) = \frac{\sum a_i b_i}{\sqrt{\sum a_i^2} \times \sqrt{\sum b_i^2}}
         .\label{eq9}\end{equation}   
        $A$, $B$: embedding vectors of two words or sentences.\\
        $a_i$, $b_i$: corresponding elements in the vector representation

\subsection{Selection Criteria for These 4 Metrics}
Evaluating medical text summarization is challenging as both accuracy and meaning retention are critical. Since summaries can be either extractive (retaining exact words) or abstractive (paraphrasing), a combination of lexical (word-matching) and semantic (meaning-based) metrics is used for a balanced evaluation.

\begin{table}
\caption{\textbf{Evaluation Metrics Comparison for Text Summarization Models}}
\label{table}
\setlength{\tabcolsep}{3pt}
\begin{tabular}{|p{40pt}|p{55pt}|p{65pt}|p{55pt}|}
\hline
Metric&
Measures &

Measures &

Weakness \\
\hline
BLEU &
Exact word overlap ( precision)&
Ensures critical \par medical terms are preserved& 
Fail on \par paraphrasing\\
 ROUGE-L &
Recall, sentence fluency&
Ensures summaries retain key details& 
Penalizes synonym\\
BERT Score &
Sematic meaning&
Measures true \par meaning retention& 
Computational expensive \\
SpaCy \par Similarity  &
General sentence coherence&
Lightweight semantic check& 
Less powerful \par than BERT Score \\

\hline
\end{tabular}
\label{tab1}
\end{table}

The Table \ref{tab1} overview the strength and weakness of each metric used for the evaluation. 
Combining lexical and semantic metrics ensures that the models not only retain important medical terminology but also produce meaningful and coherent summaries. This ensures that the summarization models do not simply copy words but provide a meaningful, medically accurate summaries.  
\subsection{Evaluation}
The test dataset was divided into three input-length categories (short, medium, long) to analyze how well the model performs on varying text complexities. Longer inputs introduce challenges in compressing multiple findings without loss of critical details, while shorter inputs demand concise extraction of meaning without unnecessary additions.
Since medical summaries vary in length and complexity, evaluating performance across different input sizes ensures that the model remains effective in real-world use. Below is the analysis of the evaluation results across key lexical and semantic metrics for each model.
\subsubsection{M-Passage}
The M-Passage model summarizes structured medical passages, such as radiology reports or clinical notes, which contain formal descriptions, findings, and diagnoses. These texts are already well-structured, meaning the model needs to condense them while preserving key medical information.\\
    \indent \textbf{Long inputs (110–141 words)}: These contain detailed passages with multiple medical findings (e.g., pneumothorax, fractures, nodules). The complexity makes summarization challenging, as the model must condense multiple findings while preserving meaning.\\
    \indent  \textbf{Medium inputs (55–75 words)}: These are moderate-length reports with fewer critical findings (e.g., atelectasis, granulomas). Summaries should retain key medical information without unnecessary details.\\
    \indent \textbf{Short inputs (20–24 words)}: These are brief reports with minimal findings, requiring the model to extract concise yet meaningful summaries.

\begin{table}[H]
\centering
\caption{\textbf{Performance Metrics of M-Passage Model}}
\label{table}
\setlength{\tabcolsep}{3pt}
\begin{tabular}{|p{60pt}|p{40pt}|}
\hline
Metric&

Avg Score \\
\hline
BLEU &
0.0982 \\
 ROUGE-L &

0.3728\\
BERT Score &
0.6533 \\
SpaCy Similarity  &
0.8413  \\

\hline
\end{tabular}
\label{tab:passage table}
\end{table}

The Table \ref{tab:passage table} represents the average score across the samples, with key observations including: \\
   \indent \textbf{Low lexical match (BLEU: 0.0982 \& ROUGE-L: 0.3728)}: The model does not rely on direct word overlap from the original text, indicating that it paraphrases instead of copying. While this can improve readability, it also raises the risk of semantic drift (a word meaning change over time), which may change the intended meaning of the medical text.\\
    \indent \textbf{Moderate semantic match (BERTScore: 0.6533)}:  The BERTScore indicates that the summaries closely align in meaning with the original text, even when phrasing differs. This is expected since medical reports often use technical language that the model must compress without altering medical accuracy.\\
    \indent  \textbf{High overall similarity (SpaCy Similarity: 0.8413)}: The high score indicate the effectiveness of the summaries when it comes to retaining the original meaning while rephrased. 

M-Passage performs well at extracting key medical details while preserving important terminology and structure. The lexical scores (BLEU, ROUGE-L) indicate that the model partially relies on paraphrasing but more on directly retaining key medical terms. Meanwhile, the strong semantic similarity scores (BERTScore, SpaCy Similarity) confirm that even when the wording changes, the essential medical meaning is preserved.
\subsubsection{M-Conversation}
The M-Conversation model processes dialogue-based medical interactions, where conversations between patients and doctors contain redundant exchanges, clarifications, and informal phrasing. The goal of this model is to extract the core medical information while omitting unnecessary conversational elements.\\
    \indent  \textbf{Long inputs (2992 - 3050 words)}: These contain detailed exchanges with extensive symptom descriptions, family history, and with many form of redundant information. The size of the inputs pose a challenge to the model, as it must remove unnecessary information then retain the key medical details. \\
    \indent \textbf{Medium inputs (1186 - 1373 words)}: Less excessive information, with moderate detail and actionable advice. \\
    \indent  \textbf{Short inputs (628 - 818 words)}: These are quick conversations with concise conversations with quick diagnoses and simple recommendations.

\begin{table}[h!]
\centering
\caption{\textbf{Performance Metrics of M-Conversation Model}}
\label{table}
\setlength{\tabcolsep}{3pt}
\begin{tabular}{|p{60pt}|p{40pt}|}
\hline
Metric&

Avg Score \\
\hline
BLEU &
0.0018 \\
 ROUGE-L &

0.0848\\
BERT Score &
0.4707 \\
SpaCy Similarity  &
0.8801  \\

\hline
\end{tabular}
\label{tab3}
\end{table}

It can easily be seen that the lexical scores (BLEU \& ROUGE-L) are significantly lower than that of M-Passage as shown in Table \ref{tab3}. Below are the further observations: \\
    \indent \textbf{Very low lexical match (BLEU: 0.0018 \& ROUGE-L: 0.0848)}:  The extremely low BLEU and ROUGE-L scores indicate that the generated summaries use different words and phrasing from the original input. This is expected because M-Conversation does not aim to retain exact words but rather extracts relevant medical details, discarding fillers like greetings, confirmations, or repetitive clarifications.

    \indent  \textbf{Mediocre semantic match (BERTScore: 0.4707)}:  The BERTScore shows that while the generated summaries do not match the input lexically, they still capture some of the intended meaning. The lower score compared to other models suggests that some nuances from the conversation may be lost or altered, which is a challenge in conversational summarization. The score reflects that while key information is retained, some phrasing differences may affect the model’s ability to capture all details perfectly.
    
   \indent \textbf{High overall similarity (SpaCy Similarity: 0.8801)}: The high SpaCy Similarity score indicates that, despite significant rewording, the overall meaning between the input conversation and the generated summary remains well-preserved. Since this metric evaluates sentence embeddings rather than exact word overlap, it confirms that the model effectively compresses conversational exchanges into meaningful medical insights without losing context.

M-Conversation performs well in extracting core medical details but does so at the cost of lexical overlap, leading to low BLEU and ROUGE-L scores. However, the strong semantic similarity suggests that the model successfully condenses the conversation into a medically relevant summary, aligning with its intended purpose.
\subsubsection{M-Question}
The M-Question model is designed to extract the key question from a given medical text. Unlike M-Passage and M-Conversation, which summarize larger sections of text, M-Question focuses on identifying and isolating the core inquiry within the input. This often involves significant paraphrasing and restructuring of sentences, which affects its evaluation scores.

    \indent  \textbf{Long inputs (134–179 words)}: These inputs detailed personal health descriptions with specific question.\\
    \indent  \textbf{Medium inputs ( 47–53 words)}: These are the balance between some background health information and the specific question.\\
    \indent  \textbf{Short inputs (9–19 words)}: These are the concise, direct question.

\begin{table}[H]
\centering
\caption{\textbf{Performance Metrics of M-Question Model}}
\label{table}
\setlength{\tabcolsep}{3pt}
\begin{tabular}{|p{60pt}|p{40pt}|}
\hline
Metric&

Avg Score \\
\hline
BLEU &
0.0084 \\
 ROUGE-L &

0.1938\\
BERT Score &
0.5552 \\
SpaCy Similarity  &
0.8630  \\

\hline
\end{tabular}
\label{tab4}
\end{table}
\begin{figure*}[h]
    \centerline{\includegraphics[scale=.5]{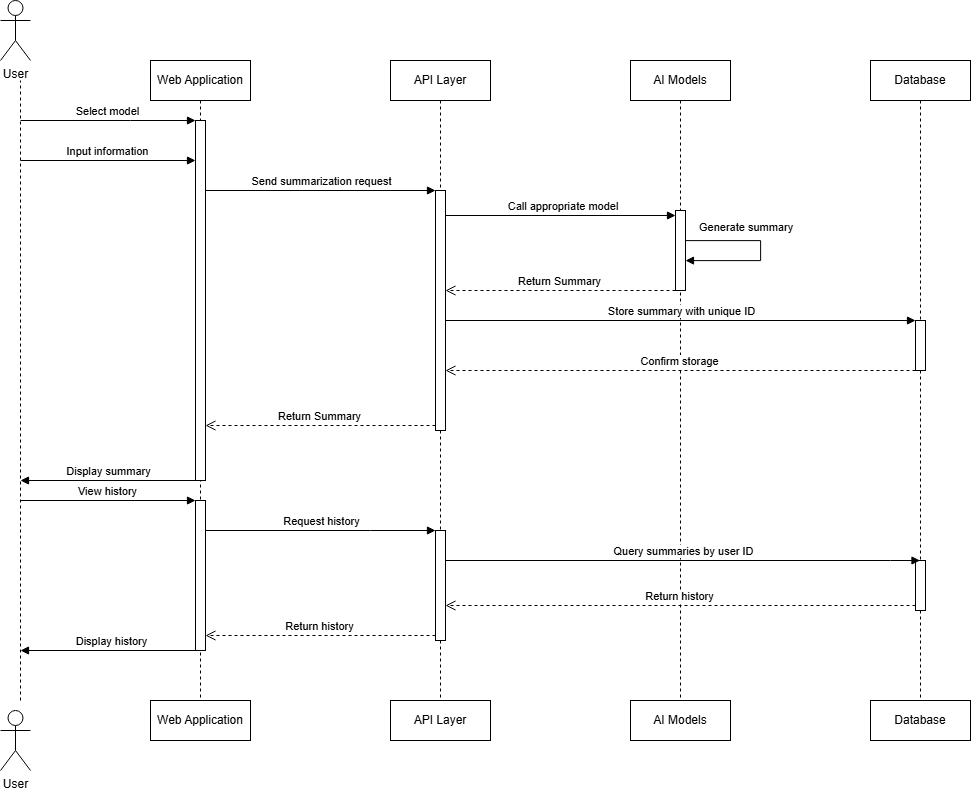}}
    \caption{Sequence Diagram for Text Summarization Process}
    \label{fig:text sum}
\end{figure*}

The Table \ref{tab4} represents the M-Question metric score. The significant outcomes as follows: 

    \indent \textbf{Very low lexical match (BLEU: 0.0084, ROUGE-L: 0.1938)}: BLEU is close to zero, indicating that the generated question rarely uses exact word sequences from the input text. ROUGE-L, while higher than BLEU, is still low, suggesting that only partial phrase matches occur between the generated question and the original passage. This is expected because questions are often restructured or reworded significantly, rather than directly copying phrases from the input.
   
    \indent  \textbf{Moderate semantic match (BERTScore: 0.5552)}: The BERTScore shows that, despite the low word overlap, the generated questions retain the core meaning of the original inquiry. This aligns with the purpose of M-Question, which is to extract and rephrase the main question only from complex medical text rather than just shortening it.

   \indent  \textbf{High overall similarity (SpaCy Similarity: 0.8630)}: The strong SpaCy Similarity score suggests that the generated questions successfully capture the intent of the original query. This result is rather surprisingly high as it only captures the main question while leaving non-question context out of the summaries. The score shows an effective pre-train process.
 
M-Question achieves its goal of isolating medical inquiries, but it does so primarily through paraphrasing rather than direct word matching. This explains the very low BLEU and ROUGE-L scores, while the higher BERTScore and SpaCy Similarity indicate strong semantic retention. The model successfully restructures medical questions, ensuring that they remain concise, clear, and contextually relevant without relying on exact word repetition.

\subsection{Summary of the Evaluation Results}

The evaluation of the three models: M-Passage, M-Conversation, and M-Question revealed distinct performance characteristics based on their intended tasks. While all models effectively condensed medical information, their scores varied depending on the nature of the input and the summarization objectives.\\
  \indent \textbf{Lexical vs. Semantic performance}: A common pattern across all models was the contrast between low lexical match (BLEU, ROUGE-L) and strong semantic retention (BERTScore, SpaCy Similarity). This suggests that the models rely heavily on paraphrasing rather than direct word overlap. While it fulfill its purpose as a summarization model, it also introduces potential risks of semantic drift, meaning word loses meaning over time, especially in medical contexts where accuracy is crucial.\\
  \indent \textbf{Model specific observations}: M-Passage achieved a moderate balance between lexical and semantic scores, indicating that it effectively condenses structured medical reports while preserving key medical terminology. The moderate BERTScore (0.6533) and high SpaCy Similarity (0.8413) confirm that its summaries retain meaning, though occasional paraphrasing may alter finer details. M-Conversation exhibited the lowest lexical match (BLEU: 0.0018, ROUGE-L: 0.0848) due to its need to filter out conversational exchanges. Despite this, it maintained the highest SpaCy Similarity (0.8801), confirming that its summaries accurately captured the core medical details. However, its moderate BERTScore (0.4707) suggests that some nuances from doctor-patient dialogues were lost in the transformation.M-Question displayed the lowest BLEU score (0.0084) due to significant rewording required to extract key medical inquiries. However, its BERTScore (0.5552) and high SpaCy Similarity (0.8630) indicate that while lexical overlap was minimal, the extracted questions remained faithful to their original intent.
  
These insights provide a foundation for further improvements, such as refining paraphrasing techniques or incorporating additional constraints to maintain medical accuracy. Overall, the models demonstrate strong summarization capabilities, with each satisfy its purpose in their specific domain.
\section{Application developments}
The development of the Medalyze application integrates AI models into a functional system. This section describes how the AI models were packaged, how the database was structured, and the design of the web interface. The goal is to ensure efficient model deployment and an intuitive user experience.
The sequence diagram (Fig. \ref{fig:text sum}) depicts the interaction between the user, web application, API layer, AI models, and database during the summarization process and history retrieval. The workflow consists of the following steps:\\
\textbf{ Summarization process:}
    \begin{itemize}
        \item The user selects the desired model and inputs medical-related text in the web application.
        \item The web application forwards a summarization request to the API layer.
        \item The API layer calls the appropriate AI model to process the input.
        \item The AI model generates a summary and sends it back to the API layer.
        \item The API layer stores the summary in the database with a unique ID.
        \item The database confirms successful storage.
        \item The processed summary is returned to the web application, where it is displayed to the user.
    \end{itemize}
   \indent\textbf{ History retrieval process:}
    \begin{itemize}
        \item The user requests their summary history.
        \item The web application sends a request to the API layer.
        \item The API layer queries the database for summaries linked to the user ID.
        \item The database retrieves and returns the user’s summary history.
        \item The history is displayed in the web application.
    \end{itemize}
\subsection{Packaging the AI models}
After training, the best variation of each model is saved as \textbf{model.safetensors}. This format was chosen for its optimized memory usage, with a touch of security benefit. Additionally, \textbf{T5Tokenizer} was imported and paired with each model to generate tokens, which will be used to make the communication with the web front-end possible. 
To make the trained models accessible, Flask, a lightweight web framework, was used to create API endpoints. The models were loaded using the Hugging Face \textbf{transformers} library, which simplifies the process of handling tokenization.
\subsection{Back-end Development and API Integration}
The API and back-end serve as the central system connecting the AI models, database, and front-end interface. This component handles requests, processes data, and ensures that summaries generated by the AI models are properly stored and retrieved when needed.\\
\subsubsection{API Design and Implementation}
A Flask-based REST API was created to allow communication between the front end, AI models, and database. The API exposes several endpoints, including:

    \indent \textbf{ Summarization endpoint:} Receives text input, processes it through the AI model, and returns the summarized result.\\
     \indent \textbf{ Database interaction:} Handles requests for storing and retrieving summaries from the database.\\
     \indent \textbf{ Health check:} Provides system status updates to monitor API availability.

The API runs within a Docker container, ensuring a structured environment that includes all necessary dependencies. This allows for stable performance regardless of where the application is deployed.

\subsubsection{Back-end and Database Communication}
The back-end is responsible for managing the flow of data between different components by directing data between the AI model, API, and database, ensuring summaries are correctly processed and stored. It interacts with \textbf{YugabyteDB}, the chosen database for Medalyze which will be explained in the following section, through standard PostgreSQL drivers, allowing efficient execution of queries for inserting, fetching, and managing summary data. Each summary is stored with a UUID-based primary key, making it easy to track and retrieve specific records.
To maintain a structured workflow, API endpoints were assigned specific tasks, reducing redundancy and improving maintainability. By following this approach, the system ensures that requests are processed effectively while maintaining clear separation between different functions.
\subsection{Database Design}
\begin{figure*}[h]
    \centerline{\includegraphics[scale=.35]{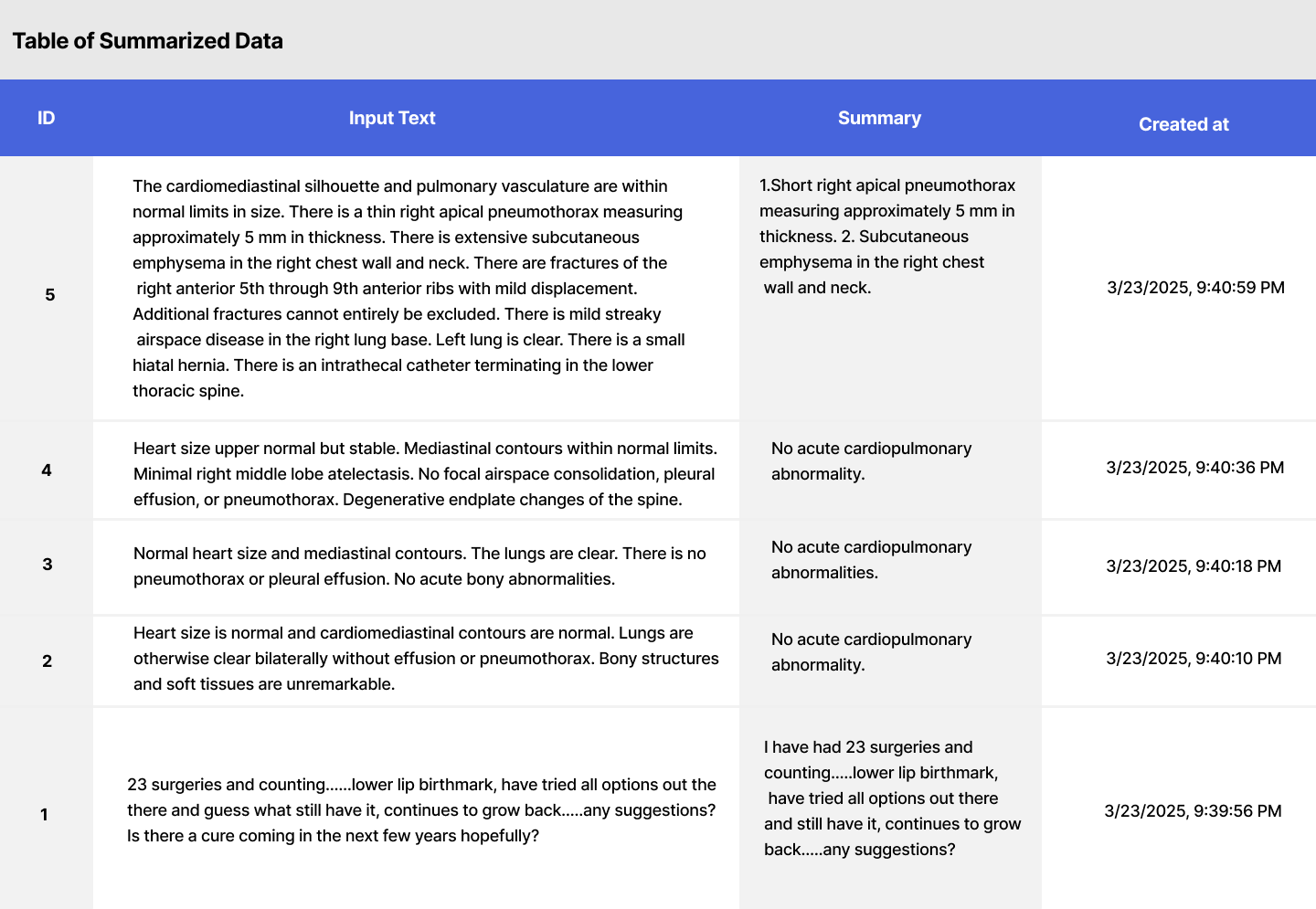}}
    \caption{User History Section Interface of Medalyze}
    \label{fig:history}
\end{figure*}
The database is responsible for storing generated summaries, ensuring they can be accessed and retrieved efficiently. As mentioned in the previous section, \textbf{YugabyteDB} was chosen as the database system. It was selected as the database due to its compatibility with PostgreSQL and its ability to handle structured queries efficiently.
Although YugabyteDB is typically used in distributed environments, Medalyze runs a single-node instance, making it function similarly to a standard relational database. This choice ensures reliable data storage while keeping an option for possible future enhancement.
The database consists of a single table, \textbf{Summaries}, which includes:

    \indent\textbf{ ID (UUID, Primary key):} A unique identifier for each summary entry.\\
    \indent\textbf{ Input (TEXT):} The original text provided by the user.\\
    \indent\textbf{ Summarized (TEXT):} The processed summary output.\\
    \indent\textbf{ Created\_time (TIMESTAMP):} The date and time when the summary was generated.

This design ensures that stored summaries can be retrieved quickly and in an organized manner. Using a UUID prevents conflicts in identification, and the timestamp field allows sorting entries chronologically.
\subsection{Front-end Development}
The web application serves as the primary interface for users to interact with the summarization models. The platform allows users to input their medical text (or text file), select a summarization model, and view the generated summaries in real-time. Additionally, it provides a history feature where users can review past summarization results.
The web application features a clean and user-friendly interface, with a front-end framework that connects to the back-end and AI models through efficient API communication. This section covers the design choices and use-case study of Medalyze. 
\subsubsection{Design}
The design of the web application prioritizes simplicity and efficiency, ensuring users can interact with the system seamlessly. This section covers the structural layout, key UI elements, and technology choices that shape the application’s functionality. 
Before implementation, the interface was first prototyped in Figma to establish a clear layout and user flow. Once the design was finalized, the front end was developed using JavaScript, HTML, and CSS, while Node.js was used to handle interactions between the interface and the API layer.\\
The final web interface was designed for quick and efficient use, eliminating unnecessary elements like login screens. Users can instantly input medical text and receive summaries without additional steps. The main UI layout consists of:\\
   \indent\textbf{ Text input area:} A large space for typing or pasting medical text, with an option of uploading text file (.txt). \\
    \indent\textbf{ Summarization output:} Displaying generated summaries.
    A history section (Fig. \ref{fig:history}) provides users access to previous summaries. Each past summary entry includes:\\
     \indent \textbf{ Summarization ID:} An automated ID is attached to each summarization.\\
       \indent \textbf{ Input Text:} Shown past summarization inputs.\\
     \indent\textbf{ Summary:} The generated summarized content.\\
     \indent\textbf{ Created At:} The exact date and time the summary was created.

\subsubsection{Use case diagram}
The use case diagram (Fig. \ref{fig:use-case}) illustrates the primary interaction between the user and the web interface. It defines how users interact with different summarization functionalities and the system’s responses, including:
\begin{figure}[tb]
    \centering
    \includegraphics[width=0.9\linewidth]{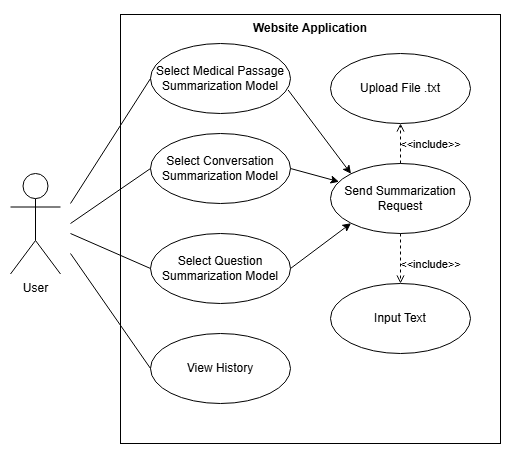}
    \caption{Use-Case Diagram for Medalyze System}
    \label{fig:use-case}
\end{figure}

         \indent\textbf{ Model selection:} Users can choose from three summarization models.
        This selection determines which AI model will process the input text.
      
         \indent\textbf{ Submitting a summarization request:}
             Users provide text input or upload a .txt file for processing.
        
         \indent\textbf{ Viewing summarization history:}
            Users can retrieve previous summarization results.
             This feature provides access to input texts, summaries, and timestamps.

The use case diagram provides a structured overview of how users interact with Medalyze’s summarization features. By outlining key actions such as model selection, text submission, and history retrieval, it helps define the functional scope of the application.
\section{Comparison with GPT-4}
Evaluating the performance of the fine-tuned Flan-T5-Large models against GPT-4 provides insights into their capabilities in medical text summarization. GPT-4 is a large-scale transformer model designed for a wide range of language tasks, whereas the Flan-T5-Large models have been specifically optimized for summarizing medical text.
This section presents a structured comparison between these models, covering specifications, evaluation methodology, performance metrics, and key findings. The assessment is based on BLEU, ROUGE-L, BERTScore, and SpaCy Similarity metrics, alongside an analysis of output quality. The goal is to determine how these models perform in handling medical text and identify their respective advantages and limitations.
\subsection{Model Specifications}
This section covers the key attributes of the fine-tuned Flan-T5-Large models and GPT-4. Key attributes are highlighted in Table \ref{tab:model spec}. 

\begin{table}[tb]
\caption{\textbf{Model Specifications of Flan-T5-Large and GPT-4}}
\label{table}
\setlength{\tabcolsep}{3pt}
\begin{tabular}{|c|c|c|c|c|c|}
\hline
Model&
Context &
Parameter &
Proprietary &
Seq2seq &
Autoreg. \\

\hline
FLAN-T5-Large&
512 &
$\sim$ 780 $M$&
$-$&
yes&
\\
GPT-4 &
32.768&
unknown & 
yes & 
$-$ & 
yes
\\

\hline
\end{tabular}
\label{tab:model spec}
\end{table}

 In term of context length, Flan-T5-Large is 512 tokens, meaning it can only process a limited input sequences. While GPT-4 is 32,768 tokens, allowing for much longer text input and better context retention. Flan-T5-Large is around 780 millions parameters, making it a relatively lightweight model. GPT-4 is unknown parameter size, but estimated to be in the trillions, making it significantly larger. Flan-T5-Large is sequence-to-Sequence (Seq2Seq) model, meaning it generates an output sequence based on an input sequence, useful for summarization. GPT-4 is autoregressive, meaning it predicts the next token based on previously generated tokens (better for open-ended text generation.
Flan-T5-Large is much smaller in comparison to GPT-4 but its structure is more centralize around summarization. While GPT-4 is significantly larger and capable of handling context and meaning longer.
\subsection{Evaluation Setup}
\begin{figure*}[tb]
	\centerline{\includegraphics[scale=.54]{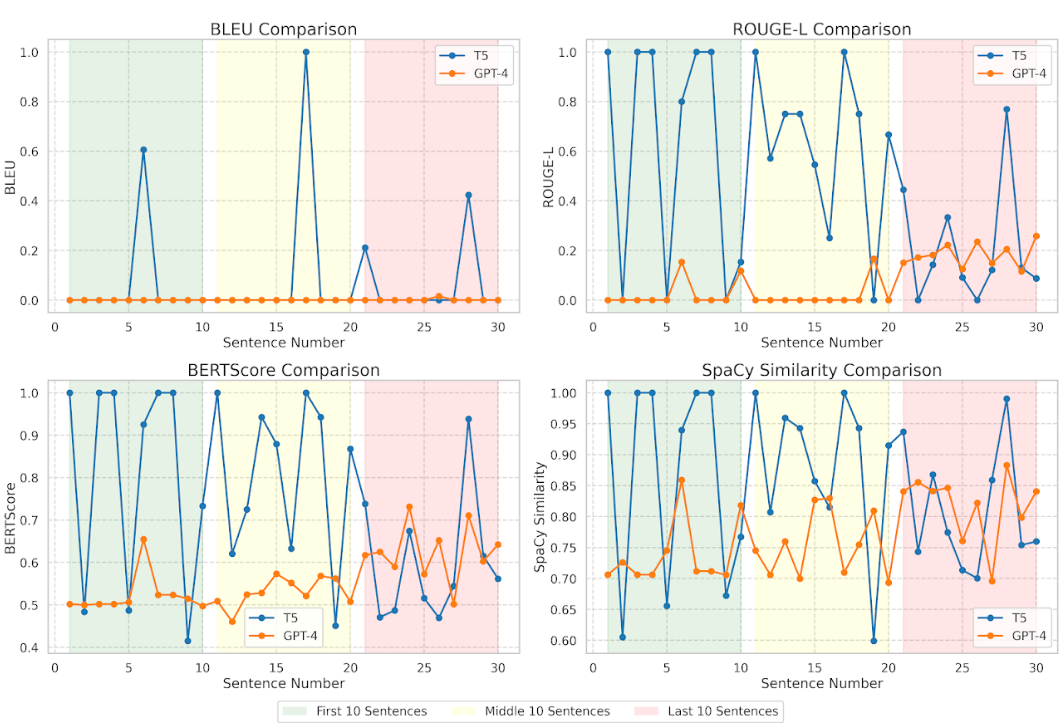}}
        \caption{Performance Comparison between M-Passage Model and GPT-4}
    \label{fig:passage gpt4}
\end{figure*}
To ensure a fair and objective comparison, both Flan-T5-Large and GPT-4 are evaluated using the same test dataset and identical evaluation metrics (BLEU, ROUGE-L, BERTScore, and SpaCy-based assessments). This guarantees that performance differences arise solely from the models’ capabilities rather than dataset variations.
A key consideration in this comparison is the \textbf{verification of the correctness} of generated text. Since the test dataset is derived from the original training dataset, the human-written target summaries serve as the ground truth. This ensures that the generated outputs are assessed against a well-defined and accurate reference point. 
Additionally, to better visualize the evaluation results, each Flan-T5-Large model will be compared to GPT-4 individually. Along with it are: 
\begin{itemize}
    \item \textbf{A table}: side by side view of the average score of each model among
the four metrics.
    \item \textbf{A figure}: an in-depth, sample by sample result of each model among the four metrics. The blue line signify Flan-T5-Large's result while the orange line is for the GPT-4. With colored box for each 10 samples represent the length of input (green = short input, yellow = medium input, red = long input).
\end{itemize}
\subsection{Performance Comparison}
The effectiveness of the summarization models is assessed using standardized evaluation metrics, including BLEU, ROUGE-L, BERTScore, and SpaCy similarity. This section compares the performance of the fine-tuned Flan-T5-Large models with GPT-4 across three summarization tasks: medical passage summarization, extracting health-related information from a conversation, and pinpointing the main medical question.
To ensure a fair comparison, GPT-4 is tested on the same dataset and evaluated using identical metrics. The following subsection provides an overview of how GPT-4 was adapted to perform these tasks before presenting the performance results.
\subsubsection{GPT-4 Summarization Tasks and Scores}
Since GPT-4 is being compared with three models, it has to be given a clear instruction that will results in the same form of summarization. 
\begin{figure*}[tb]
	\centerline{\includegraphics[scale=.54]{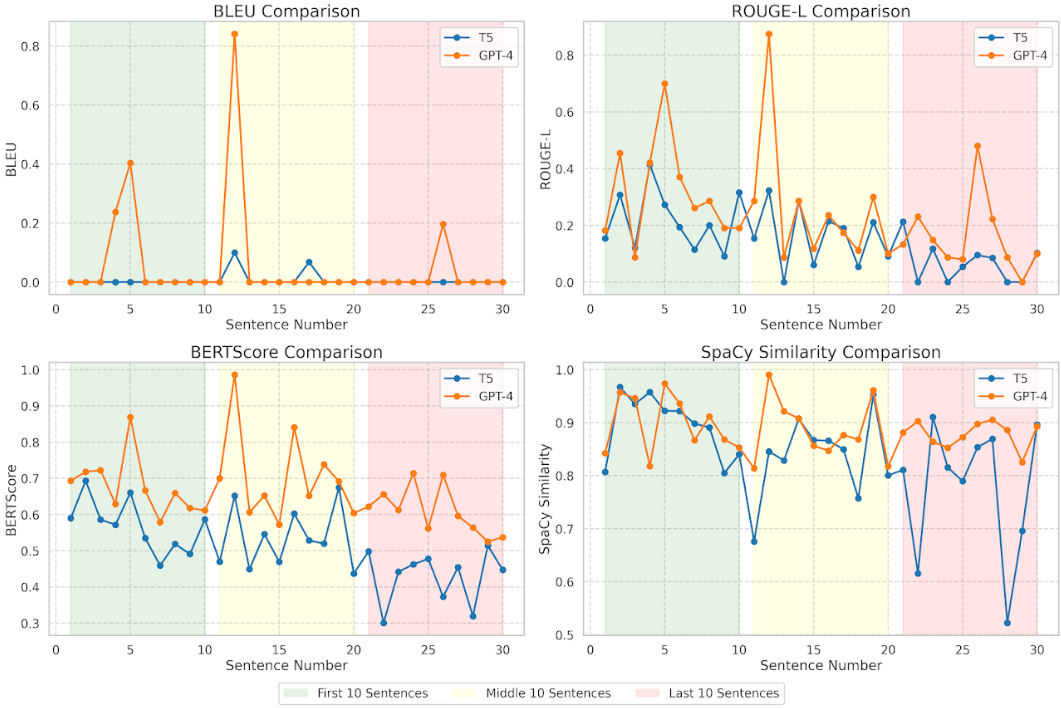}}
    \caption{Performance Comparison between M-Question Model and GPT-4}
    \label{fig:question gpt4}
\end{figure*}
\begin{figure*}[tb]
	\centerline{\includegraphics[scale=.54]{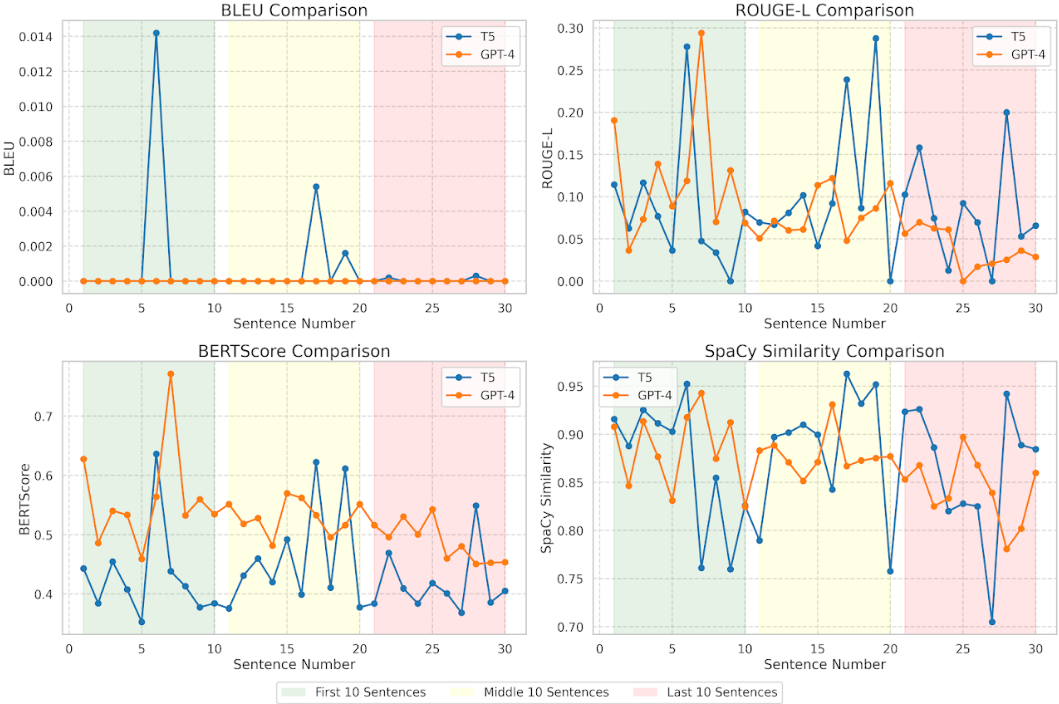}}
    \caption{Performance Comparison between M-Conversation Model and GPT-4}
    \label{fig:conversation gpt}
\end{figure*}

\begin{table}[H]
    \centering
      \caption{Average Evaluation Scores of GPT-4 on Summarization Tasks}
    \begin{tabular}{|c|c|c|c|}
        \hline
         Metric    & Passage & Question & Conversation  \\
         \hline
         BLEU    &  0.0032 & 0.0911 & 0.0000 \\
        
         ROUGE-L & 0.0644 & 0.2564 & 0.0779 \\
         
         BERTScore & 0.5453 & 0.6524 & 0.5428 \\
         
         SpaCy Similarity & 0.7810 & 0.8879 & 0.8682\\
         \hline
    \end{tabular}
  
    \label{tab:gpt label}
\end{table}
Table \ref{tab:gpt label} presents the average evaluation scores of GPT-4 across the three summarization tasks.
\subsubsection{M-Passage vs GPT-4}
This section evaluates the fine-tuned M-Passage model performance against GPT-4 on medical passage summarization.
\begin{table}[H]
    \centering
    \caption{Comparison of M-Passage Model and GPT-4 on Summarization Tasks}
    \begin{tabular}{|c|c|c|}
    \hline
      Metric   & M-Passage & GPT-4 \\
      \hline
      BLEU   & 0.0982 & 0.0032\\
      ROUGE-L & 0.3728 & 0.0644 \\
      BERTScore & 0.6533 & 0.5453 \\
      SpaCy Similarity & 0.8413 & 0.7810\\
      \hline
    \end{tabular}

    \label{tab:passage vs gpt}
\end{table}
Based on the Table \ref{tab:passage vs gpt}, key observations are: 
\begin{itemize}
    \item The largest gaps are in BLEU and ROUGE-L, suggesting that M-Passage’s outputs have more word overlap with reference summaries.
    \item BERTScore and SpaCy Similarity favor M-Passage, meaning it better retains semantic meaning.
\end{itemize}
While difference in lexical match does not indicate whether a model is better than the other, M-Passage achieved the higher semantic match means its summaries manage to convey the medical information more effectively than GPT-4.

Based on the Fig. \ref{fig:passage gpt4}, key observations are as follow:
\begin{itemize}
    \item M-Passage has high variation, with peaks and dips in its scores.
    \item BLEU \& ROUGE-L: M-Passage dominates but fluctuates, showing it sometimes generates highly accurate summaries and sometimes struggles.
    \item BERTScore \& SpaCy Similarity: M-Passage maintains a stronger semantic match, but its instability suggests sensitivity to different passage complexities not in the scope of length.
\end{itemize}
While the Table \ref{tab:passage vs gpt} shows a higher semantic match for M-Passage, the Fig. \ref{fig:passage gpt4} shows inconsistencies in generated summaries. In the field of summarization, it is important to be consistent with the output quality, even more in medical domain. A potential improvement could involve stabilizing M-Passage’s performance across different text segments while maintaining its accuracy advantages.

\subsubsection{M-Question vs GPT-4}
This section evaluates the fine-tuned M-Question model performance against GPT-4 on primary medical question identification. 
\begin{table}[tb]
    \centering
      \caption{Comparison of M-Question Model and GPT-4 on Question Extraction}
    \begin{tabular}{|c|c|c|}
    \hline
      Metric   & M-Question & GPT-4 \\
      \hline
      BLEU   & 0.0084 & 0.0911\\
      ROUGE-L & 0.1938 & 0.2564 \\
      BERTScore & 0.5552 & 0.6524 \\
      SpaCy Similarity & 0.8630 & 0.8879\\
      \hline
    \end{tabular}
  
    \label{tab:question gpt4}
\end{table}
Based on the Table \ref{tab:question gpt4}, key observations are: 
\begin{itemize}
    \item The largest margin of improvement is in BLEU (0.0911 vs. 0.0084) and ROUGE-L (0.2564 vs. 0.1938), indicating better word overlap and phrase-level matching for GPT-4.
    \item BERTScore and SpaCy Similarity are higher for GPT-4, showing better semantic understanding.
\end{itemize}
Based on the properties of this task, being finding the main question in a passage, the output will not cover much of the original text. Meaning lower lexical score might indicate a greater focus on the main question. However, this does not discredit GPT-4 by any mean. For longer inputs with more complex phrasing and questioning, GPT-4 will point out the question but give slight details, which will effect its lexical performance.

Based on the Fig. \ref{fig:question gpt4}, key observations are: 
\begin{itemize}
    \item Besides BLEU, both models show a lot of fluctuation across all type of input. 
    \item Overall, GPT-4 performance is superior across all metrics. 
\end{itemize}
With the addition of the Fig. \ref{fig:question gpt4}, it is clear that GPT-4 had a superior performance. This result is partially due to the model type. GPT-4's autoregressive nature allows it to generate more structured and question-like outputs, while Flan-T5-Large, being a seq2seq model, suffers from fluency, especially for longer input, due to the nature of this task. 

\subsubsection{M-Conversation vs GPT-4}
This section evaluates the fine-tuned M-Conversation model performance against GPT-4 on extracting health related issues in a conversation.
\begin{table}[tb]
    \centering
    \caption{Comparison of M-Conversation Model and GPT-4 on Conversational Summarization}
    \begin{tabular}{|c|c|c|}
    \hline
      Metric   & M-Conversation & GPT-4 \\
       \hline
      BLEU   & 0.0018 &  0.0000\\
      ROUGE-L & 0.0848 & 0.0779 \\
      BERTScore & 0.4707 & 0.5428 \\
      SpaCy Similarity & 0.8801 & 0.8682\\
      \hline
    \end{tabular}

    \label{tab:conversation gpt4}
\end{table}
Based on the Table \ref{tab:conversation gpt4}, key observations are: 
\begin{itemize}
    \item BLEU: M-Conversation achieves a slightly higher score (0.0018) than GPT-4 (0.0000), suggesting minimal n-gram overlap.
    \item ROUGE-L: M-Conversation (0.0848) and GPT-4 (0.0779) perform similarly, indicating comparable recall of key phrases.
    \item BERTScore: GPT-4 (0.5428) outperforms M-Conversation (0.4707), suggesting better semantic alignment with reference texts.
    \item SpaCy Similarity: M-Conversation (0.8801) slightly surpasses GPT-4 (0.8682), implying that its outputs maintain closer linguistic similarity.
\end{itemize}
The comparison between M-Conversation and GPT-4 reveals notable differences in their summarization performance. While both models exhibit minimal n-gram overlap, as indicated by their near-zero BLEU scores, M-Conversation slightly outperforms GPT-4 in ROUGE-L and SpaCy Similarity, suggesting that its summaries retain more key phrases and structural consistency with reference texts. However, GPT-4 achieves a higher BERTScore, indicating stronger semantic alignment and a better grasp of contextual meaning. These results suggest that M-Conversation prioritizes structural approach, whereas GPT-4 generates more semantically coherent responses.

The primary observations derived from Fig. \ref{fig:conversation gpt} are listed below:
\begin{itemize}
    \item BLEU Comparison: BLEU scores remain close to zero, reinforcing that n-gram-based similarity is very low, practically none.
    \item ROUGE-L Comparison: Shows fluctuations, with M-Conversation slightly higher in certain short and medium-length conversations.
    \item BERTScore Comparison: GPT-4 maintains a more stable performance across all sentence lengths.
    \item SpaCy Similarity: Both models perform similarly, with M-Conversation being higher slightly in some regions.
\end{itemize}
Evaluating the performance of this task is inherently difficult. Unlike traditional summarization, extraction-based summarization condenses information without fully reinterpreting it making it less suited for conversational inputs. As noted in the model evaluation section, the input for this task can reach up to 3050 words, primarily consisting of conversational exchanges, with the health-related portion being relatively minimal. Consequently, lexical-based metrics, especially BLEU, which relies on n-gram matching, yield very low scores. This reflects the challenge of capturing key medical information from lengthy and dialogue-heavy texts.
\section{Conclusion}
This paper detailed the development of Medalyze, covering the entire workflow from acquiring and preparing medical text datasets to fine-tuning models and integrating them into a functional web application. As a local application, Medalyze ensures user data privacy and offline accessibility, designed for ease of use, Medalyze allows users to summarize medical texts without technical barriers. 

The comparison with GPT-4 served as a reference point to assess Medalyze’s performance, highlighting its strengths and areas for improvement in summarizing medical text. The evaluation provided valuable insights into how well the fine-tuned models handle domain-specific summarization compared to a general-purpose AI system like GPT-4.

It is important to clarify that Medalyze is not intended for medical diagnosis or prediction; its sole function is to summarize medical information for easier understanding. It acts as a supportive tool to enhance information accessibility, helping users process complex medical texts more efficiently.

\section*{Acknowledgment}
The ASEAN IVO (\url{http://www.nict.go.jp/en/asean_ivo/index.html}) project, “Artificial Intelligence Powered Comprehensive Cyber-Security for Smart Healthcare Systems (AIPOSH)”, was involved in the production of the contents of this publication and financially supported by NICT (\url{http://www.nict.go.jp/en/index.html}).

\begin{IEEEbiography}[{\includegraphics[width=1in,height=1.25in,clip,keepaspectratio]{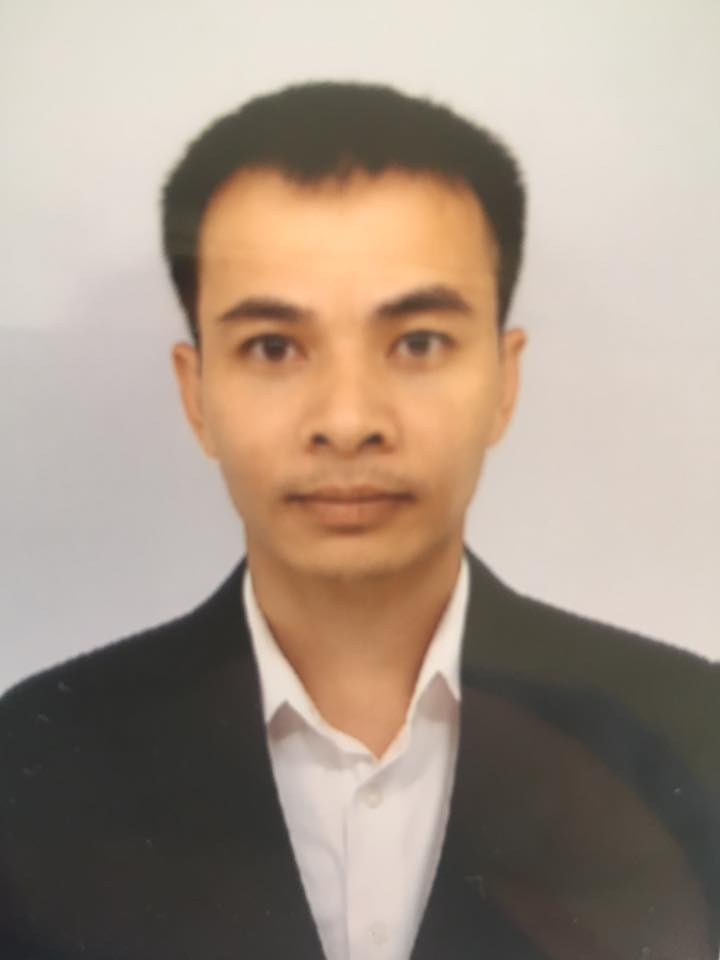}}]{Van-Tinh Nguyen} received the B.S. degree in Radio electronics engineering from
the Belarusian State University of Informatics and Radioelectronics, in 2012 and the Ph.D. degree in
computer science from Division of Information Science, Nara Institute of Science and Technology, Nara, Japan, in 2022.
 Since 2023, he has been a lecturer in the
Electrical Engineering Department of Le Quy Don Technical University, Ha Noi, Viet Nam.
His research interests include machine learning, bigdata, blockchain, and hardware security.
\end{IEEEbiography}

\begin{IEEEbiography}[{\includegraphics[width=1in,height=1.25in,clip,keepaspectratio]{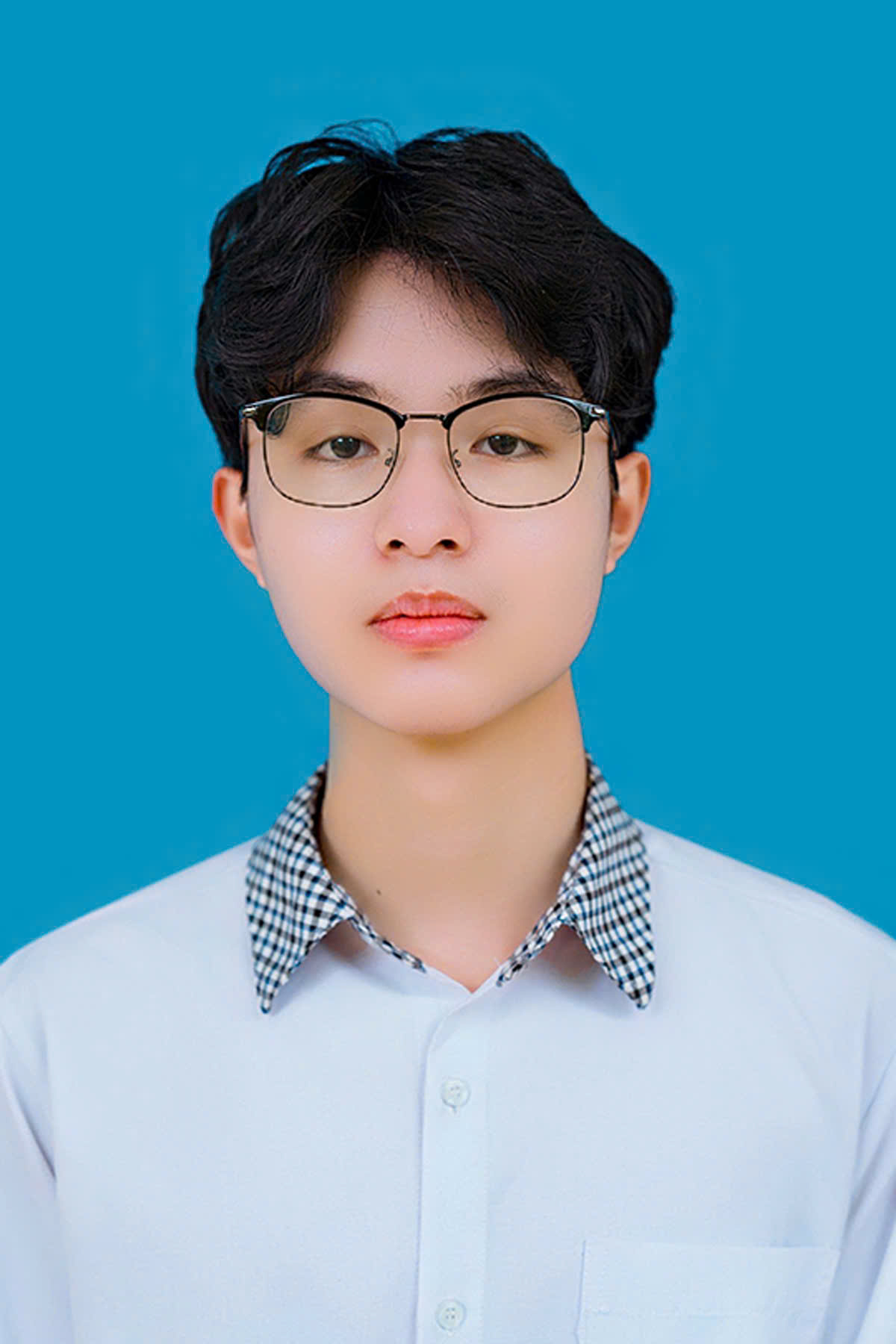}}]{Hoang-Duong Pham}  is currently a senior student majoring in Information and Communication Technology at the University of Science and Technology of Hanoi (USTH), Vietnam. He is expected to graduate in 2025. His interests include software architecture, and machine learning. 
\end{IEEEbiography}

\begin{IEEEbiography}[{\includegraphics[width=1in,height=1.25in,clip,keepaspectratio]{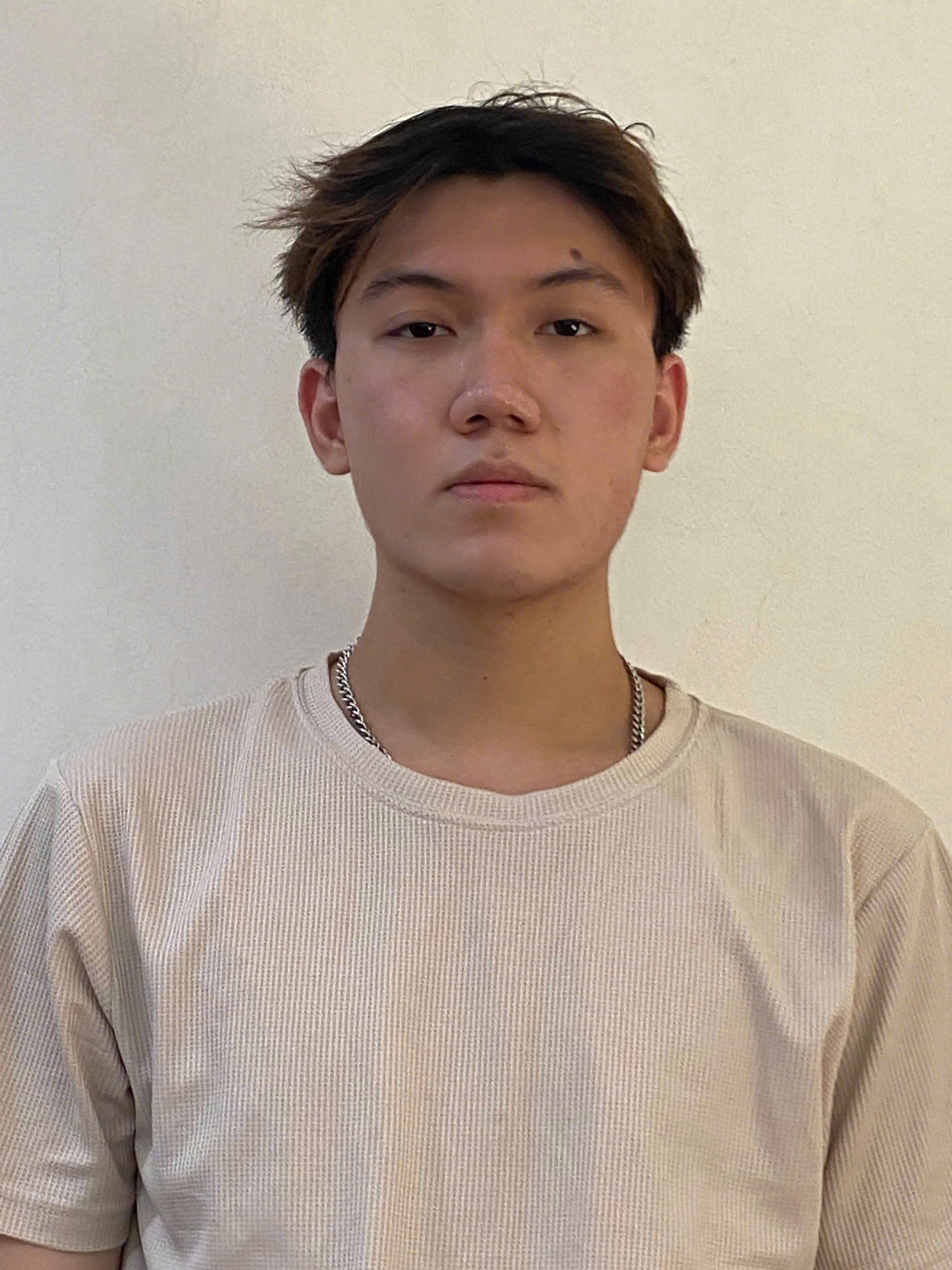}}]{Thanh-Hai To}  is currently a senior student majoring in Information and Communication Technology at the University of Science and Technology of Hanoi (USTH), Vietnam. He is expected to graduate in 2025. His interests include software architecture, blockchain, and machine learning. 
\end{IEEEbiography}

\begin{IEEEbiography}[{\includegraphics[width=1in,height=1.25in,clip,keepaspectratio]{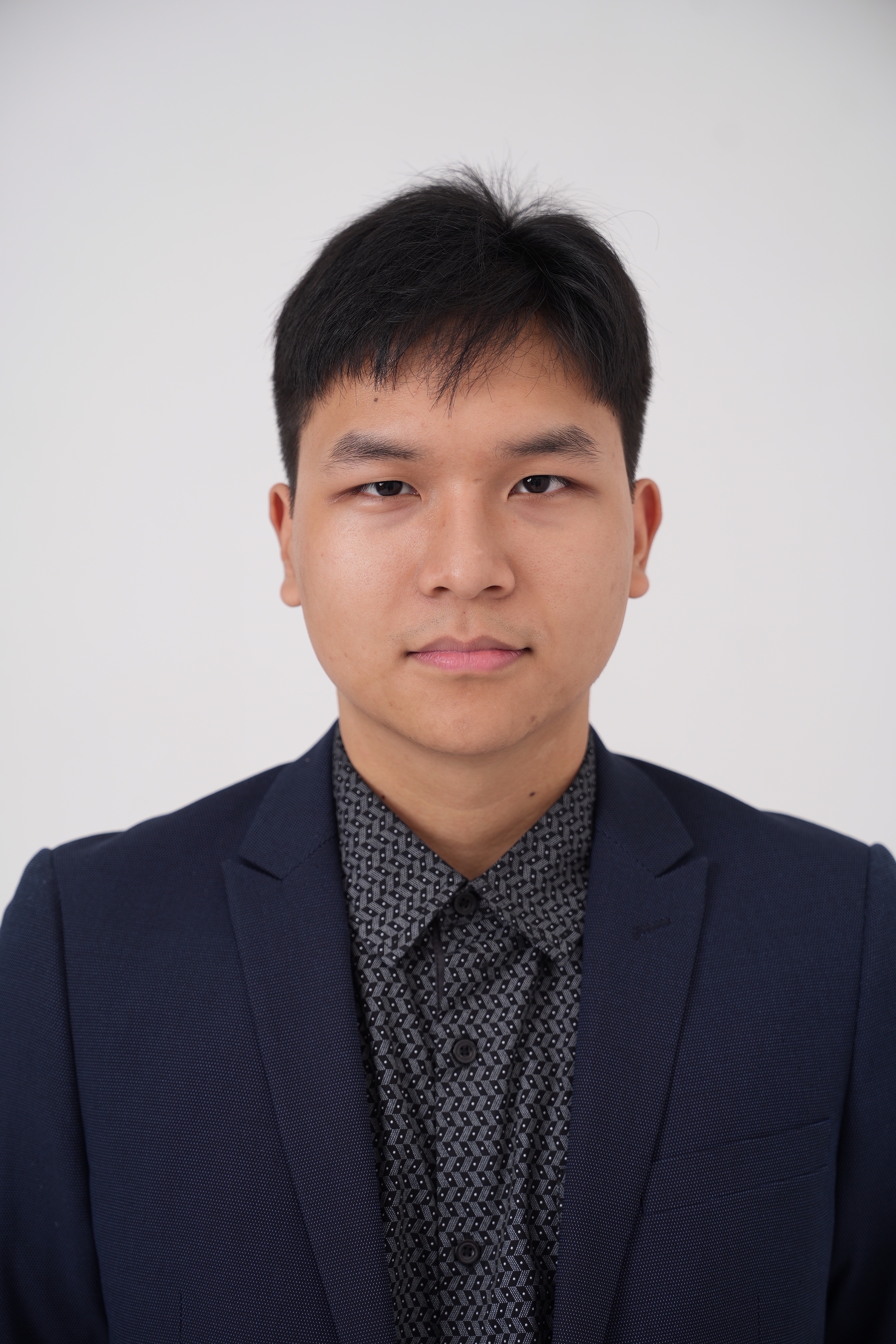}}]{Hung Do Cong Tuan} is currently a senior student majoring in Information and Communication Technology at the University of Science and Technology of Hanoi (USTH), Vietnam. He is expected to graduate in 2025. His interests include big data, workflow engines, and machine learning. 
\end{IEEEbiography}

\begin{IEEEbiography}[{\includegraphics[width=1in,height=1.25in,clip,keepaspectratio]{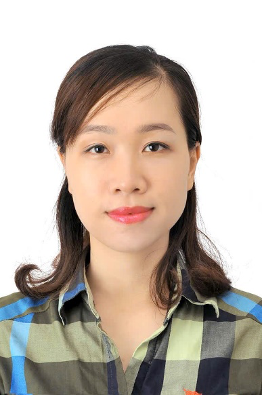}}]{Thi-Thu-Trang Dong} graduated from General Medicine in 2009 at Viet Nam Military Medical Academy - Ha Noi - Viet Nam. After 3 years, she completed her master's program, level I specialist and resident doctor at Viet Nam Military Medical Academy - Ha Noi - Viet Nam. In 2022, she graduated with a PhD degree in Neuroscience at Nagasaki University - Nagasaki - Japan. From 2023 to present, she is a lecturer and treating physician at the Department of acute diseases and Emergency - Institute for the treatment of senior Staff, 108 Institute of Clinical Medical and Pharmaceutical Sciences, Hanoi, Vietnam.
\end{IEEEbiography}

\begin{IEEEbiography}[{\includegraphics[width=1in,height=1.25in,clip,keepaspectratio]{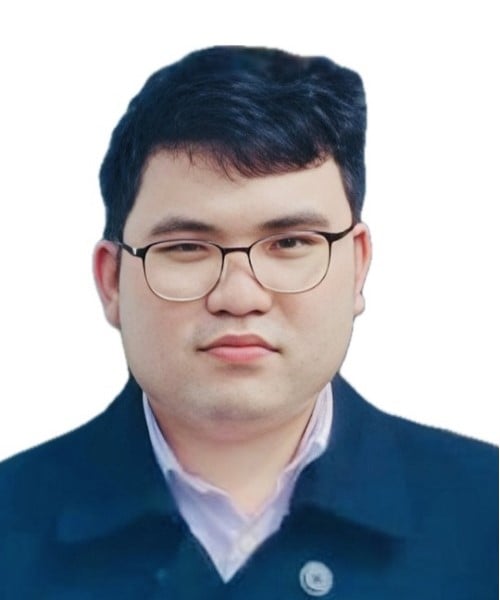}}]{Vu Trung Duong Le} received the Bachelor of Engineering degree in IC and hardware design from Vietnam National University Ho Chi Minh City (VNU-HCM)—University of Information Technology (UIT) in 2020, and the Master’s degree in information science from the Nara Institute of Science and Technology (NAIST), Japan, in 2022. He completed his Ph.D. degree in 2024 at NAIST, where he now serves as an Assistant Professor at the Computing Architecture Laboratory. His research interests include computing architecture, reconfigurable processors, and high-efficiency accelerators for quantum emulators, cryptography, and artificial intelligence.
\end{IEEEbiography}


\begin{IEEEbiography}[{\includegraphics[width=1in,height=1.25in,clip,keepaspectratio]{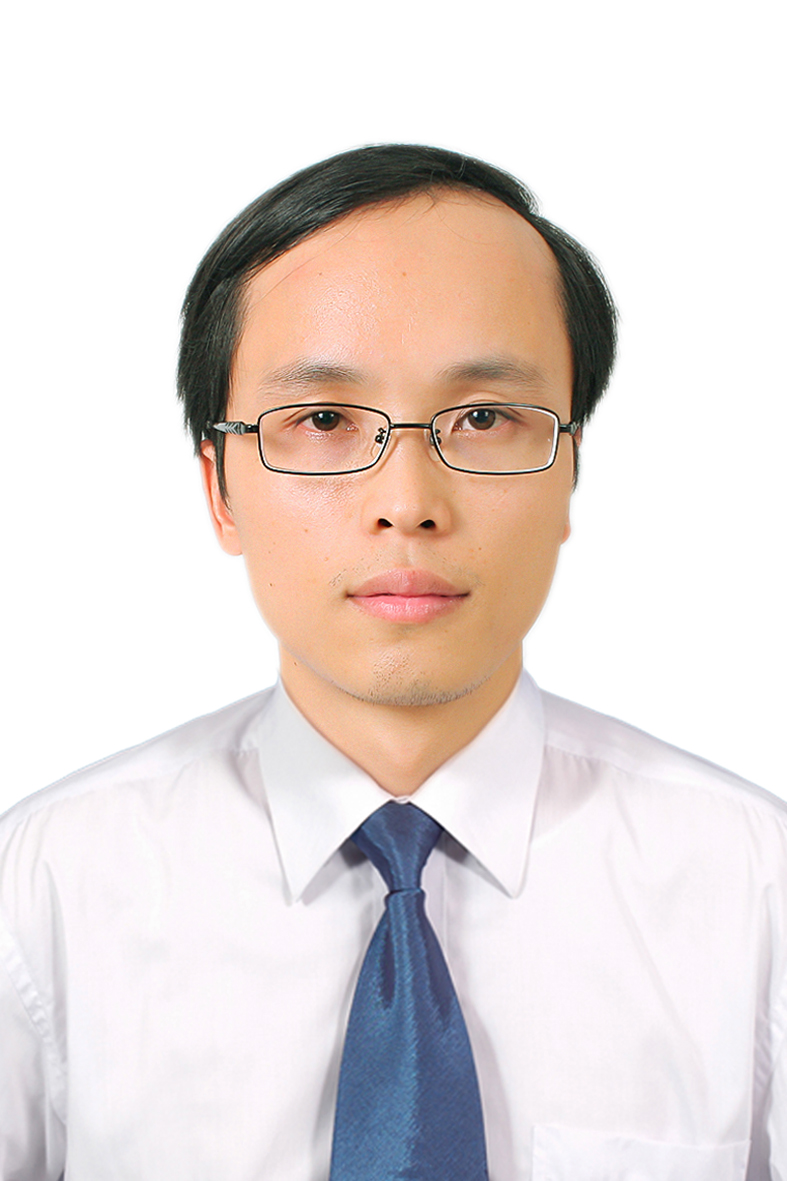}}]{Van-Phuc Hoang} received PhD degree in Electronic Engineering from The University of Electro-Communications, Tokyo, Japan in 2012. He has worked as postdoc researcher, visiting scholar at The University of Electro-Communications, Tokyo, Japan, Telecom Paris, France and University of Strathclyde, Glasgow, UK during the period of 2012-2018. He is working as an Associate Professor, Director with Institute of System Integration, Le Quy Don Technical University, Hanoi, Vietnam. His research interests include hardware security, VLSI design, digital signal processing and intelligent systems for Internet of Things. He was the Technical Program Chair of several IEEE international conferences such as ICDV 2017, MCSoC 2018, SigTelCom 2019, APCCAS 2020 and ATC 2020, ICICDT 2022, ICICDT 2023.
\end{IEEEbiography}

\EOD

\end{document}